\documentclass[runningheads]{llncs}
\usepackage{tabularx}
\usepackage{graphicx}
\usepackage{lineno,hyperref}
\modulolinenumbers[5]
 
\usepackage{url}
\usepackage{amsmath}
\usepackage{amsfonts}
\usepackage{multirow}
\usepackage[caption = false]{subfig}
\usepackage{algorithm} 
\usepackage{algpseudocode}
\usepackage{setspace}
\usepackage{multirow}
\usepackage[table]{xcolor}
\usepackage{hhline}
\usepackage{soul}
\usepackage{booktabs}
\usepackage{xcolor}
\usepackage{marvosym}
\newcommand{\yh}[1]{\textcolor{black}{#1}}
\newcommand{\xy}[1]{\textcolor{black}{#1}}

\makeatletter
\newcommand*{\rom}[1]{\expandafter\@slowromancap\romannumeral #1@}
\makeatother
\begin{document}

\newcolumntype{L}[1]{>{\raggedright\arraybackslash}p{#1}}
\newcolumntype{C}[1]{>{\centering\arraybackslash}p{#1}}
\newcolumntype{R}[1]{>{\raggedleft\arraybackslash}p{#1}}
\title{Representation Learning for Short Text Clustering}
%
\author{
Hui Yin\inst{1}\and
Xiangyu Song\inst{1}\and 
Shuiqiao Yang\inst{2}\and
Guangyan Huang\inst{1}\and \\
Jianxin Li\inst{1} \textsuperscript{\Letter}
}
%
%
\institute{
School of Information Technology, Deakin University, Geelong, Australia \\
\email{\{yinhui, xiangyu.song, guangyan.huang, jianxin.li\}@deakin.edu.au} \and
Data Science Institute, University of Technology Sydney, Sydeny, Australia
\email{shuiqiao.yang@uts.edu.au}\\}
\maketitle              
\begin{abstract}

Effective representation learning is critical for short text clustering due to the sparse, high-dimensional and noise attributes of short text corpus.
Existing pre-trained  models (e.g., Word2vec and BERT) have greatly improved the  expressiveness for short text representations with more condensed, low-dimensional and continuous features compared to the traditional Bag-of-Words (BoW) model.
However, these models are trained for general purposes and thus are suboptimal for the short text clustering task.
In this paper, we propose two methods to exploit the unsupervised autoencoder (AE) framework to further tune the short text representations based on these pre-trained text models for optimal clustering performance.
In our first method Structural Text Network Graph Autoencoder (STN-GAE), we exploit the structural text information among the corpus by constructing a text network, and then adopt graph convolutional network as encoder to fuse the structural features  with the pre-trained text features for text representation learning.
In our second method Soft Cluster Assignment Autoencoder (SCA-AE), we adopt an extra soft cluster assignment constraint on the latent space of autoencoder to encourage the learned text representations to be more clustering-friendly.
We tested two methods on seven popular short text datasets, and the experimental results show that when only using the pre-trained model for short text clustering, BERT performs better than BoW and Word2vec.
However, as long as we further tune the pre-trained representations,
the proposed method like SCA-AE can greatly increase the clustering performance, and the accuracy improvement compared to use BERT alone could reach as much as 14\%.

\end{abstract}

\section{Introduction}
Nowadays, the popularity of social media and online forums (e.g., Twitter, Yahoo, StackOverflow, Weibo) have led to the generation of millions of texts each day. 
The form of texts such as micro-blogs, snippets, news titles, and question/answer pairs have become essential information carriers \cite{cai2020target,li2020community,chen2019contextual,haldar2019location}. 
These texts are usually short, sparse, high-dimensional, and semantically diverse \cite{yang2020short}.
Short text analysis has wide applications,
 such as grouping similar documents (news, tweets, etc.), question-answering systems, event discovery, document organization, and spam detection \cite{yang2019discovering,yin2020detecting}.
Furthermore, a lot of other work also revolves around social media data \cite{kong2019copfun,wang2020efficient,tian2019evidence,jiang2019sentence}.

Text representation learning is important, and lots of recent studies have demonstrated that effective representations can improve the performance of text clustering \cite{xie2016unsupervised,jiang2016variational,aljalbout2018clustering}.
Traditionally, texts are represented by the Bag-of-Words (BoW) model, where binary weights or term frequency–inverse document frequency (TF-IDF) weights are used for text representation.
\yh{Later, pre-trained models such as Word2vec \cite{mikolov2013efficient}, GloVe \cite{pennington2014glove} are used to generate word embedding}.
In addition to obtaining the word vector, the embedding also reflects the spatial position of the word, i.e., words with a common context (semantic similarity) are mapped close to each other in the latent space with similar embeddings.
Recently, Bidirectional Encoder Representations from Transformers (BERT) \cite{devlin2018bert} model has caused a sensation in the machine learning and natural language processing communities.
It learns the text representation with transformer, an attention mechanism that can learn the contextual relations between words (or sub-words).
BERT can understand different embeddings for the same word according to the context.
Specifically, if the word represents different meanings in different sentences, the word embedding is also different.
For example, ``Apple is a kind of fruit." and ``I bought the latest Apple."
BERT can produce different embedding for ``Apple" in different contexts, while word2vec and glove only generate one embedding.

Above advanced pre-trained models can generate effective text representations, and general clustering models, such as $K$-means, are applied to these text representations to implement text clustering tasks.
\xy{The more effective the text representation is, the more accurate $K$-means can classify text into different clusters based on the similarity between text embeddings.}
\xy{Although pre-trained models significantly improve text representation ability compared to the traditional BoW model by incorporating semantic contexts into text representations.}
However, these models are generally trained on a large scale of text corpus and focused on general-purpose text mining tasks and thus are sub-optimal for the short text clustering purpose.
Therefore, we are highly motivated to fine tune the text representations generated by the pre-trained text models like BERT to further improve the performance of short text clustering. 

Specifically, we propose to exploit the unsupervised autoencoder framework with different neural network modules to further compress the text representations generated by the pre-trained models into more condensed representations.
In our first method, named STN-GAE, we propose to combine the text representations from the pre-trained models and the structural text network to achieve more condensed text representation for short text clustering. 
To achieve this, we first construct a text network where each node represents a text, and the edge between nodes indicates that the two texts have semantic similarities. 
The pre-trained text models return the features of the nodes.
Then, we exploit 
graph convolutional networks (GCN) into the autoencoder framework as an encoder to fuse the structure-level and feature-level information for condensed text representation learning. 
In our second method, named SCA-AE, we exploit a soft cluster assignment constraint \cite{jiang2016variational} into the embedding space of autoencoder for clustering-friendly feature learning.
The soft cluster assignment constraint assumes that the latent embeddings are generated from different components. 
Firstly, a soft cluster assignment distribution $P$ for each data is calculated based on Student's t-distribution \cite{van2008visualizing}, where the embeddings close to the component centers are highly confident. 
Then, an induced distribution $Q$ is further calculated based on $P$, and a KL-divergence \cite{kullback1951information} loss between $P$ and $Q$ is adopted to increase the confidence of highly confident embeddings.
Based on the two proposed methods, we can further tune the text representations provided by the pre-trained models to improve the short text clustering performance.
\renewcommand{\labelenumi}{\Roman{enumi}.}

In summary, we highlight our contributions as follows:
\begin{itemize}
   \item [\textbullet] We compare the performance of two popular pre-trained text models (Word2vec and BERT) on short text clustering tasks and find that BERT based pre-trained models could provide better representations than Word2vec for short text clustering tasks.
   \item [\textbullet] We propose two methods (STN-GAE and SCA-AE) to fine tune the pre-trained text representations and improve the short text clustering performance.
   STN-GAE adopts graph convolutional networks to combine text networks and text features for representation learning.
    SCA-AE  exploits a soft cluster assignment constraint into autoencoder to learn clustering-friendly representations.  
    
    \item [\textbullet] Extensive experiments on seven real-world short text datasets have demonstrated that with fine-tuned text representations based on the pre-trained text models, we can achieve much better clustering performance than using the pre-trained model alone. 
   \xy{The proposed SCA-AE can improve the clustering accuracy by  $14\%$.}
   \end{itemize}

The rest of this paper is organized as follows.
Section \ref{related work} reviews the related work.
Section \ref{methodology} details the framework and methodology for short text clustering.
Section \ref{experiment} presents the experimental studies. Conclusions are made in Section \ref{conclusion}.


\section{Related Work}
\label{related work}
In this section, we survey three techniques for short text clustering, e.g., text representation learning, text structure information extracting, and the soft cluster assignment for clustering purposes.
\subsection{Text Representation Learning}
Traditionally, Bag-of-Words (BoW) is the simplest way to generate a text vector, where the dimension of the vectors is equal to the size of the vocabulary.
While for short text, the vector tends to contain many empty values, resulting in sparse and high-dimensional vectors. 
Also, there is no information about the sentence syntax or the order of words in the text. 
Word2vec and GloVe are popular unsupervised learning algorithms for learning representations of words, and they are context-independent. 
These models output just one vector or embedding for each word in the vocabulary and combine all different senses of the word in the corpus into one vector.
Word vectors are positioned in the vector space such that words with common contexts in the corpus are located close to \xy{each other} in the space.
Pre-trained Word2vec/GloVe embeddings are publicly available, and word embeddings can also be trained based on \yh{special} domain corpus.
The text embedding is usually represented by the sum or average value of all word embeddings in the text, and the \xy{dimensionality} of the text vector is equal to the word embedding.
For the missing words in the pre-trained word embedding, ignoring unknown words or providing them with random vectors is better.


\subsection{Text Structure Information Extracting}
Some studies such as PTE \cite{tang2015pte}, Text GCN \cite{yao2019graph}, RTC \cite{liang2020robust} and SDCN \cite{bo2020structural} adopt the text structural information to supplement the text representation. 
\xy{The corpus is represented as a network where each node is a text or word and edges are created between nodes according to certain principles, so that each node can obtain additional information from neighboring nodes to enrich its representation.}
Text GCN \cite{yao2019graph} and Robust Text Clustering (RTC) \cite{liang2020robust} use the same strategy to construct a text graph that contains word nodes and document/text/sentence nodes.
For the document/text/sentence-word edges, the weight of the edge is the term frequency-inverse document frequency (TF-IDF) of the word in the document, and for the word-word edges, they use the point-wise mutual information (PMI) method to calculate the weight (only the edges with positive PMI value are preserved).
Structural Deep Clustering Network (SDCN) \cite{bo2020structural} constructs a $K$-Nearest Neighbor (KNN) graph based on the original corpus, and the samples are the nodes in the graph.
For all samples, they use heat kernel and dot-product to calculate a similarity matrix, then select \xy{the top $K$ similar neighbors for each sample} 
and set edges to connect it with them.
After building the network, 
\xy{they employ a GCN model for the network structure, and a DNN model for the text representation}, and then combine the loss function of \xy{both} models \xy{for training}.

\subsection{Soft Cluster Assignment for Text Clustering} 
The soft clustering assignment algorithm is developed using the $K$-means algorithm and its variants.
The document's assignment is a distribution over all clusters, that is, a document has fractional membership in several clusters.
Thus soft cluster assignment techniques are widely used in text clustering to improve performance.
The goal is to learn \xy{clustering-friendly} text representations, where data points are evenly distributed around the cluster centers and the boundaries between clusters are \xy{relatively clear}.
The common method includes the soft cluster assignment loss into the training objectives to optimize the learning models and learn \xy{clustering-friendly} representations. 


\section{Methodology}
\label{methodology}
In this section, we first introduce the preliminary of the autoencoder framework, then we detail two proposed methods: STN-GAE and SCA-AE.
\begin{figure}[t]
    \centering
    \includegraphics[width=115mm, scale=1.15]{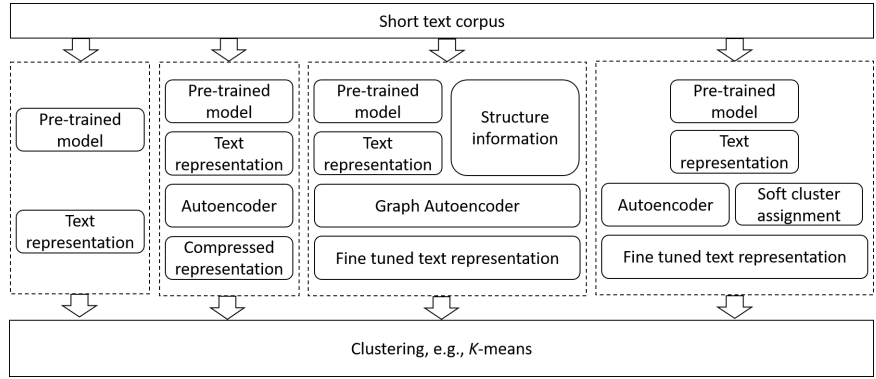}
    \caption{The general framework of this case study. \yh{The input is a short text corpus, and the output is the clustering result.}}
    \label{Framework}
\end{figure}


\subsection{Preliminary of Autoencoder}
\begin{figure}[!t]
    \centering
    \includegraphics[width=90mm, scale=0.9]{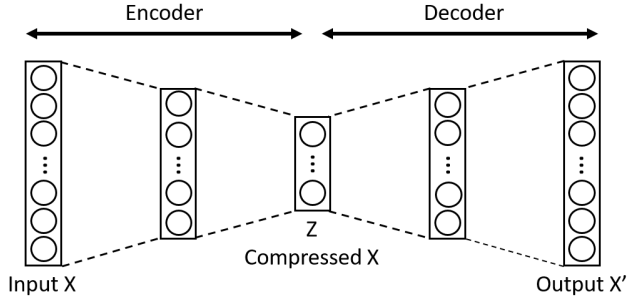}
    \caption{Autoencoder architecture. The encoder stage of the autoencoder compresses the high-dimensional input representation $X$ into a low-dimensional representation $Z$, the decoder stage maps $Z$ to a reconstruction $X^{'}$ with the same shape as $X$.}
    \label{AEStructure}
\end{figure}

We use an autoencoder to reduce the dimensionality of the features, and then directly apply $K$-means to them.
Autoencoder compresses the high-dimensional input representation $X$ into a low-dimensional representation $Z$ while retaining most original information. 
Using these compressed representations can reduce model computational complexity and running time.
Figure \ref{AEStructure} provides the architecture of the autoencoder.
We want to minimize the reconstruction loss to make reconstructed $\hat{X}$ closer to $X$ and adopt Mean Square Error (MSE) as our loss function, as shown in equation \ref{AEloss}, which measures how close $\hat{X}$ is to $X$.

\begin{equation}
\label{AEloss}
    L(X,\hat{X}) = ||X-\hat{X}||^{2}
\end{equation}

In our experiments, we perform dimensionality reduction preprocessing on BERT based text representation for each dataset and then perform $K$-means clustering.

\subsection{STN-GAE: Representation Learning with Structure Information}
\label{TextGraphConstruction}
We propose a method Structural Text Network-Graph Autoencoder (STN-GAE), which exploits the structural text information among the corpus by constructing a text network and then adopts graph convolutional network into the autoencoder (Graph Autoencoder) to fuse the structural features with the pre-trained text features for text representation learning.
This deep neural network (DNN) is able to learn non-linear mapping without manually extracting features to transform the inputs into a latent representation and maintain the original structure.
The layer-wise propagation rule of graph autoencoder \cite{kipf2016semi} is defined as follows:

\begin{equation}
    H^{(l+1)}=\sigma(\tilde{D}^{-\frac{1}{2}}\tilde{A}\tilde{D}^{-\frac{1}{2}}H^{(l)}W^{(l)})
\end{equation}

where $\tilde{A}=A+I_{N}$ is the adjacency matrix of the undirected graph $G$ with added self-connections, $I_{N}$ is the identity matrix, $D$ is the degree matrix, $\sigma$ is a non-linear activation function such as the ReLU,  $W^{(l)}$ is the weight matrix for layer $l$. 
$X$ is the high-dimensional input representation, $H^{(l)} \in R^{N \times D}$ is the matrix activation in the $l_{th}$ layer, $H^0=X$.

\begin{figure}[!t]
    \centering
    \includegraphics[width=110mm, scale=0.8]{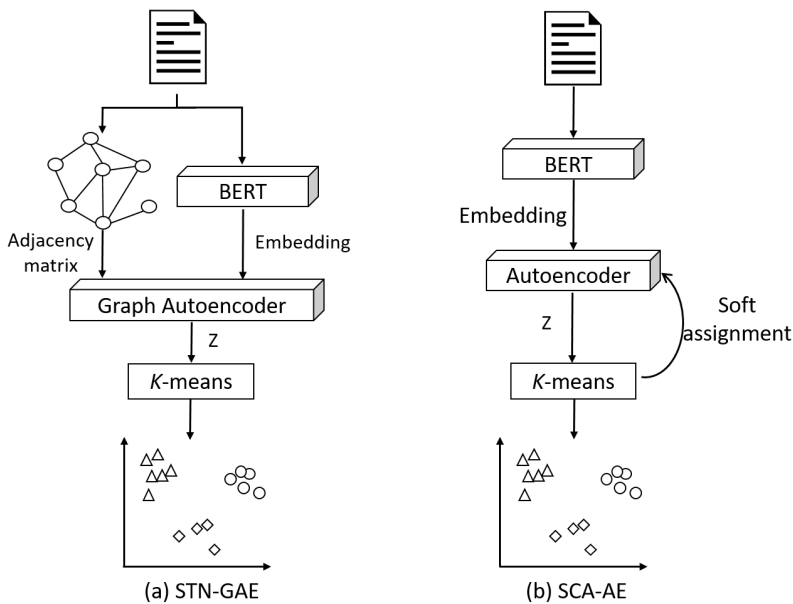}
    \caption{The structure of proposed methods STN-GAE and SCA-AE.}
    \label{StructureWhole}
\end{figure}

We first use the same strategy as SDCN \cite{bo2020structural} to construct a $K$-Nearest Neighbor (KNN) graph, and each sample is a node in the text network.
Then we input the adjacency matrix $A$ and raw features $X$ into Graph Autoencoder (GAE) and get the encoded low-dimensional embedding of the text, and then apply clustering algorithm $K$-means to them, as shown in figure \ref{StructureWhole}(a).
We will describe the strategy in detail in the following.

We calculate the cosine score between text embeddings (BERT based embeddings) and construct the semantic similarity matrix $S$ of the whole corpus.
After calculating the similarity matrix $S$, we select the top-$K$ similarity samples of each node as its neighbors to construct an undirected $K$-nearest neighbor graph. 
The above strategy ensures that each sample is connected to highly correlated $K$ nodes to ensure the integrity of the graph structure, which is essential for a convolutional neural network. 
In this way, we can get the adjacency matrix $A$ from the non-graph data.
Therefore, the text graph is represented as $G =(V, E, X)$, where $V(|V| = N)$ are the set of nodes and $E(|E| = N * K)$ are the set of edges, respectively, $K$ is the number of neighbors per sample. $X =[x_1,x_2,...,x_N] \in R^{N \times d}$ is node feature vectors matrix, where $x_i$ represents the feature vector of node $i$, and $d$ is the dimension of the feature vector, the adjacency matrix is $A \in R^{N \times N}$.

\subsection{SCA-AE: Representation Learning with Soft Cluster Assignment Constraint}

A lot of work has been confirmed that using deep neural networks to simultaneously learn feature representation and cluster assignment usually yields better results.
In method SCA-AE, we employ cluster assignment hardening loss to optimize the text representation.
This method includes three steps: (1) Use BERT model to generate text representation;
(2) Use autoencoder to reduce dimensionality to get compressed input embeddings;
(3) Use soft cluster assignment as an auxiliary target distribution, and jointly fine-tune the encoder weights and the clustering assignments to improve clustering performance.
The third step requires using soft assignment of data points to clusters. 

We use the Student’s t-distribution $Q$ \cite{van2008visualizing} as a kernel to measure the similarity between embedded nodes $z_i$ and centroids $u_j$.
This distribution $q$ is formulated as follows:
\begin{equation}
    q_{ij} =\frac{(1 + \parallel z_i - u_j\parallel ^{2})^{-1}}{\sum_{j^{'}} (1 + \parallel z_i - u_{j^{'}}\parallel ^{2})^{-1}}
\end{equation}

where $q_{ij}$ can be considered as the probability of assigning sample $i$ to cluster $j$, i.e., a soft assignment. $z_i$ is the $i$-th row of $Z_{(N)}$, $u_{j}$ is initialized by $K$-means on representations learned by DNN model, we treat $Q = [q_{ij}]$ as the distribution of the assignments of all samples. 

After obtaining the clustering result distribution $Q$, we further use an auxiliary target distribution $P$ \cite{xie2016unsupervised} to improve cluster purity and put more emphasis on samples assigned with high confidence so that the samples are closer to cluster centers.
Compared to the similarity score $q_{ij}$, $P$ is ``stricter", which can prevent large clusters from distorting the hidden feature space.
The probabilities $p_{ij}$ in the proposed distribution $P$ are calculated as:
\begin{equation}
    p_{ij} =\frac{q_{ij}^2/\sum_{i^{'}}q_{i^{'}j}}{\sum_{j^{'}}q_{ij^{'}}^2/\sum_{i^{'}}q_{i^{'}j^{'}}} 
\end{equation}

in which each assignment in $Q$ is squared and normalized. 

The KL-divergence \cite{kullback1951information} between the two probability distributions $P$, and $Q$ is then used as training objective, the training loss $L$ is defined as:
\begin{equation}
    L_{soft}=KL(P\parallel Q)=\sum_i\sum_jp_{ij}log\frac{p_{ij}}{q_{ij}}
\end{equation}

The above process is usually called self-training or self-supervised. 
By minimizing the KL-divergence loss between $P$ and $Q$ distributions, the target distribution $P$ can help the DNN model learn a clustering-friendly representation, i.e., \yh{making the sample closer to the centroid to which it belongs.}
The structure of method SCA-AE is shown in figure \ref{StructureWhole}(b).

\section{Experiments}
In this section, we report our experimental results regarding the adopted datasets, the compared methods, and the parameter settings.

\label{experiment}
\subsection{Datasets}
\label{DatasetDescription}
We choose seven short text datasets with different lengths and categories for experiments.
We use raw text as input without additional processing, such as stop word removal, tokenization, and stemming.
The statistics of these datasets are presented in table \ref{dataset}, and the detailed descriptions are the following:
\begin{itemize}
    \item[$\diamond$] MR\footnote{https://github.com/mnqu/PTE/tree/master/data/mr} \cite{pang2005seeing}: The MR is a movie review dataset for binary sentiment classification, containing 5,331 positive and 5,331 negative reviews.
    \item[$\diamond$] AGnews\footnote{https://www.kaggle.com/amananandrai/ag-news-classification-dataset}: The AG's news topic classification dataset contains around 127,600 English news articles, labeled with four categories. We use the test dataset for experiments, which includes 7,600 news titles.
    \item[$\diamond$] SearchSnippets \cite{phan2008learning}:
    A collection consists of Web search snippets categorized in eight different domains.
    The 8 domains are Business, Computers, Culture-Arts, Education-Science, Engineering, Health, Politics-Society, and Sports.
    \item[$\diamond$] Yahoo\footnote{https://www.kaggle.com/soumikrakshit/yahoo-answers-dataset}: The answers topic classification dataset is constructed using the ten largest main categories from Yahoo. 
    We randomly select 10,000 samples from the dataset.
    \item[$\diamond$] Tweets\footnote{http://trec.nist.gov/data/microblog.html}: 
    This dataset consists of 30,322 highly relevant tweets to 269 queries in the Text REtrieval Conference (TREC) 2011-2015 microblog track 1. 
    We choose the top ten largest categories for experiments.
    The total number of samples is 6,988, the sample size of each class is different, the maximum count is 1,628, and the minimum count is 428.
    \item[$\diamond$] StackOverflow\footnote{https://www.kaggle.com/c/predict-closed-questions-on-stack-overflow/}:
    A collection of posts is taken from the question and answer site StackOverflow and publicly available on Kaggle.com. We use the subset containing the question titles from 20 different categories.
    \item[$\diamond$] Biomedical\footnote{http://participants-area.bioasq.org/}:
    This is a snapshot of one year of PubMed data distributed by BioASQ to evaluate large-scale online biomedical semantic indexing. 
\end{itemize}
\begin{table}[]
\centering
\caption{Statistics of the text datasets.}
\begin{tabular}{c|c|c|c|c}

\toprule
\multicolumn{1}{c|}{\textbf{Dataset}} 
& \multicolumn{1}{c|}{\textbf{Categories} }& \multicolumn{1}{c|}{\textbf{Samples} }& \multicolumn{1}{c|}{\textbf{Ave Length}} & \multicolumn{1}{c}{\textbf{Vocabulary}} \\ \midrule
{MR}      & 2   &  10,662  & 21.02   & 18,196 \\ 
{AGnews}  & 4 & 7,600  & 6.76  & 16,406 \\ 
{SearchSnippets}  & 8                              & 12,340                        & 17.9                           & 30,610             \\              
{Yahoo}   & 10  &  10,000     & 34.16 & 23,710 \\ 

{Tweets}  & 10    & 6,988  & 7.28 & 5,773   \\

{StackOverflow} & 20  & 20,000 & 8.3   & 10,762 \\ 
{Biomedical}  & {20}       & {20,000}     & {12.9}        & {18,888}  \\       
\bottomrule
\end{tabular}

\label{dataset}
\end{table}

\subsection{Compared Methods and Parameter Settings}
\label{Baseline}


\begin{itemize}
    \item[$\diamond$] Word2vec: We use Google's pre-trained word embeddings and then take the average of all word embeddings in the text as the text representation. The words not present in the pre-trained words are initialized randomly, the text vector dimension is 300.
    \item [$\diamond$] BERT: We employ the sentence-transformer model \cite{reimers2019sentence} to generate text embeddings directly. 
    Since these short text datasets come from different sources, and the pre-trained BERT models are trained based on different types of datasets, to avoid the impact of BERT model selection on the performance of text clustering, we run three BERT models on all datasets and choose the best model of the three for each dataset.
    For datasets Yahoo, SearchSnippets, AGnews, StackOverflow, and Biomedical, the BERT model ``paraphrase-distilroberta-base-v1" gets the best performance. 
    For dataset MR, BERT model ``stsb-roberta-large" gets the best result, and for dataset Tweets, BERT model ``distilbert-base-nli-stsb-mean-tokens" is the best one.
    \item[$\diamond$] Autoencoder: We set the hidden layers to $d$:500:500:2000:10, which is consistent with other work \cite{guo2017improved,bo2020structural,xie2016unsupervised}, where $d$ is the dimension of the input data $X$ and 10 is the dimension of compressed $X$.  
    We tune other parameters and set the learning rate as 0.001, and the MSE is used to measure reconstruction loss after the decoder decodes the encoded embeddings.
    We also try other settings of the hidden layer and find that small changes do not affect the performance much.
    See section \ref{LayerAnalysis} for details.
    \item[$\diamond$] STN-GAE: We calculate the similarity matrix $S$ based on the text embeddings generated by the BERT model and use the same strategy in SDCN \cite{bo2020structural} to construct the $K$-Nearest Neighbor (KNN) graph, $K$=10.
    We set the hidden layer of GAE to $d$:64:32, and train them with 300 epochs for all datasets, the learning rate is set to 0.002.
    \item[$\diamond$] SCA-AE: We use the same setting as \cite{hadifar2019self}, for instance, the hidden layer size of all datasets to $d$:500:500:2000:20, 20 is the dimensionality of compressed text embedding.
    We set the batch size to 64, pre-trained the autoencoder for 15 epochs, and initialize stochastic gradient descent with a learning rate of 0.01 and a momentum value of 0.9.
\end{itemize}

We adopt the $K$-means as the clustering algorithm. Two popular metrics are exploited to evaluate the clustering performance, i.e., we employ two metrics: normalized mutual information (NMI) and clustering accuracy (ACC).



\begin{table}[p]
\centering
\caption{Clustering performance comparison. The results are averaged over five runs.}
\label{FinalResult}

\begin{tabular}{C{2.5cm}|C{1.2cm}|C{1.2cm}C{1.2cm}C{1.2cm}C{1.2cm}C{1.8cm}C{1.5cm}}
\toprule
\textbf{Dataset}                & \textbf{Metric} & \multicolumn{1}{c}{\textbf{BoW}} & \multicolumn{1}{c}{\textbf{W2V}} & \multicolumn{1}{c}{\textbf{BERT}} & \multicolumn{1}{c}{\textbf{AE}} & \multicolumn{1}{c}{\textbf{STN-GAE}} & \multicolumn{1}{c}{\textbf{SCA-AE}}  \\ \toprule

\multirow{2}{*}{MR}             & ACC             & 0.5188                            & 0.526                             & \textbf{0.7848}                    & 0.7492                           & 0.6625                            & 0.7662                              \\
                                & NMI             & 0.0037                            & 0.002                             & \textbf{0.2511}                    & 0.2013                           & 0.1676                            & 0.2198                              \\ 
\hline
\multirow{2}{*}{AGnews}         & ACC             & 0.2809                            & 0.3977                            & 0.5748                             & 0.6748                           & 0.5742                            & \textbf{0.6836}                     \\
                                & NMI             & 0.0094                            & 0.1193                            & 0.2577                             & 0.3299                           & 0.2914                            & \textbf{0.3414}                     \\ 
\hline
\multirow{2}{*}{SearchSnippets} & ACC             & 0.2459                            & 0.623                             & 0.6675                             & 0.6429                           & 0.4144                            & \textbf{0.6871}                     \\
                                & NMI             & 0.0893                            & 0.4782                            & 0.4763                             & 0.4619                           & 0.3167                            & \textbf{0.5026}                     \\ 
\hline
\multirow{2}{*}{Yahoo}          & ACC             & 0.1507                            & 0.1355                            & 0.4271                             & 0.4803                           & 0.3387                            & \textbf{0.5606}                     \\
                                & NMI             & 0.0413                            & 0.0264                            & 0.2765                             & 0.3069                           & 0.2529                            & \textbf{0.3472}                     \\ 
\hline
\multirow{2}{*}{Tweets}         & ACC             & 0.6161                            & 0.5771                            & 0.8126                             & 0.8424                           & 0.4049                            & \textbf{0.8485}                     \\
                                & NMI             & 0.7284                            & 0.6084                            & 0.867                              & 0.8875                           & 0.3546                            & \textbf{0.8919}                     \\ 
\hline
\multirow{2}{*}{StackOverflow}  & ACC             & 0.4615                            & 0.2289                            & 0.6253                             & 0.6672                           & 0.4049                            & \textbf{0.7655}                     \\
                                & NMI             & 0.5506                            & 0.1905                            & 0.5962                             & 0.6156                           & 0.4492                            & \textbf{0.6599}                     \\ 
\hline
\multirow{2}{*}{Biomedical}     & ACC             & 0.1388                            & 0.2296                            & \textbf{0.4043}                    & 0.3894                           & 0.1550                            & 0.4025                              \\
                                & NMI             & 0.0887                            & 0.2166                            & \textbf{0.3347}                    & 0.3385                           & 0.1468                            & 0.3329                              \\
\bottomrule
\end{tabular}
\end{table}
\subsection{Clustering Performance Analysis}
Table \ref{FinalResult} indicates the clustering results on seven datasets.
For the pre-trained BERT model, we adopt the best one selected from the three pre-trained BERT models, refer to section \ref{Baseline} for details.
AE, STN-GAE, and SCA-AE use the BERT model to generate text representations as input features.
We have the following observations:
\begin{itemize}
    \item [$\diamond$]In baseline methods (BoW, W2V, and BERT), BERT achieves the best results in all seven datasets for each metric. Compared to Word2vec and BoW, BERT has state-of-the-art semantics representation ability, which can map the text to a more accurate position in the latent space, which has a positive impact on clustering.
    \item [$\diamond$]AE and BERT achieve similar clustering results. Because the autoencoder retains the most important information in compressing the text embedding, the compressed vector is about the same as the input data.
    \item [$\diamond$]The method STN-GAE does not perform well as expected, even worse than the AE model and BERT model.
    The reason is that in the GAE model, the text representation contains structural information, when the structural information in the graph is not clear enough, the performance of the GCN-based method will decrease. 
    This conclusion is consistent with SDCN.
    \item [$\diamond$]The use of soft cluster assignment for deep clustering (SCA-AE) is able to achieve the best performance in 5 out of 7 datasets compared to the best baseline method BERT.
    In the AGnews, Yahoo, and StackOverflow datasets, the accuracy has increased by about 10\%, and the NMI has improved slightly.
    \item [$\diamond$]For dataset Biomedical, the performance is really poor in all methods, and the possible reason is this dataset belongs to the biology/medicine domain \cite{du2019neural,supriya2020automated,sarki2020automated}, and the existing pre-trained BERT models are all trained on a common social media dataset.
    \yh{A possible solution is to train a BERT model in this professional field.}
\end{itemize}

\begin{figure*}[p]
     \centering
     \subfloat[BoW \label{BoW}]{\includegraphics[width=0.33\textwidth]{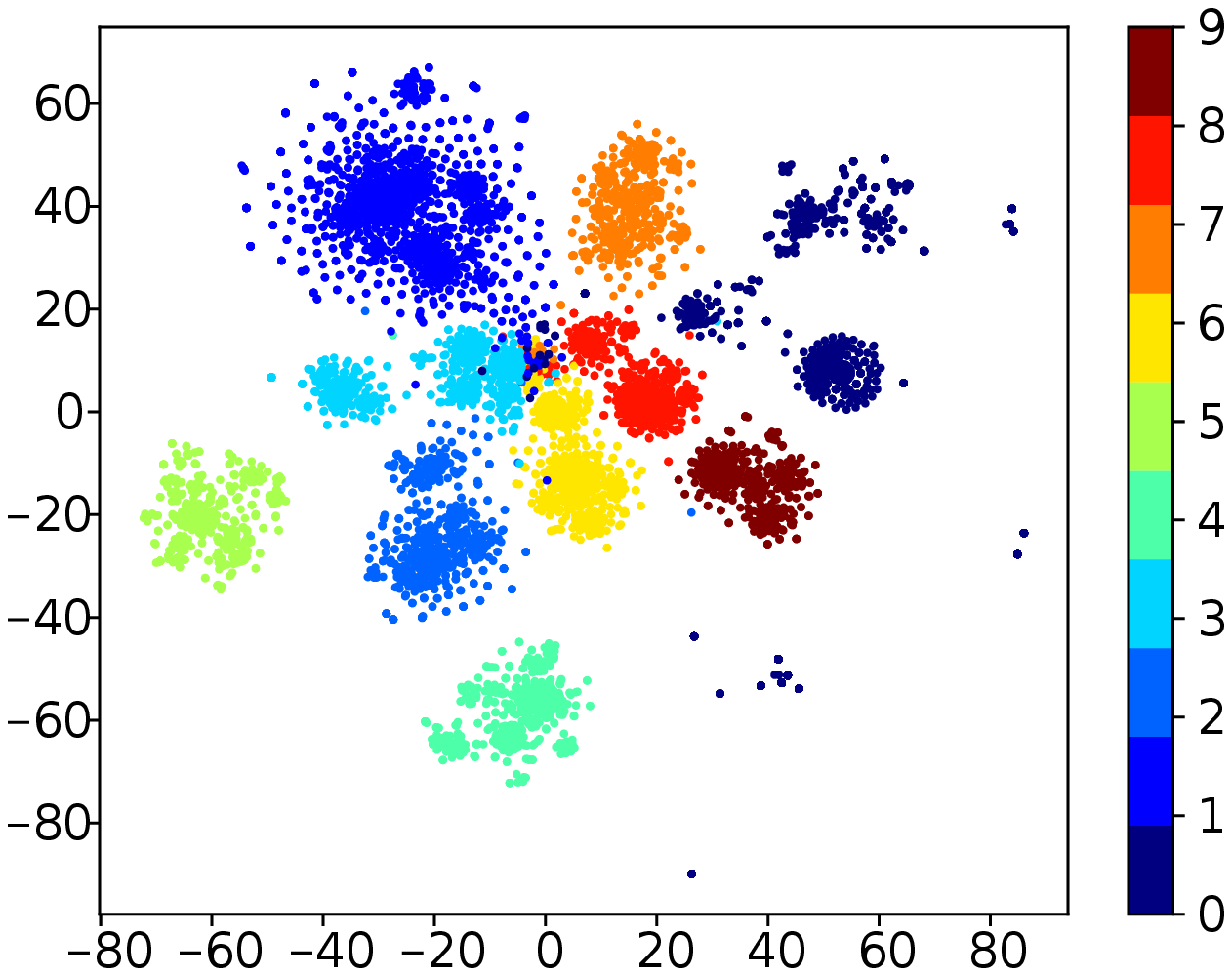}}\hfill
     \subfloat[Word2vec
     \label{Word2vec}]{\includegraphics[width=0.33\textwidth]{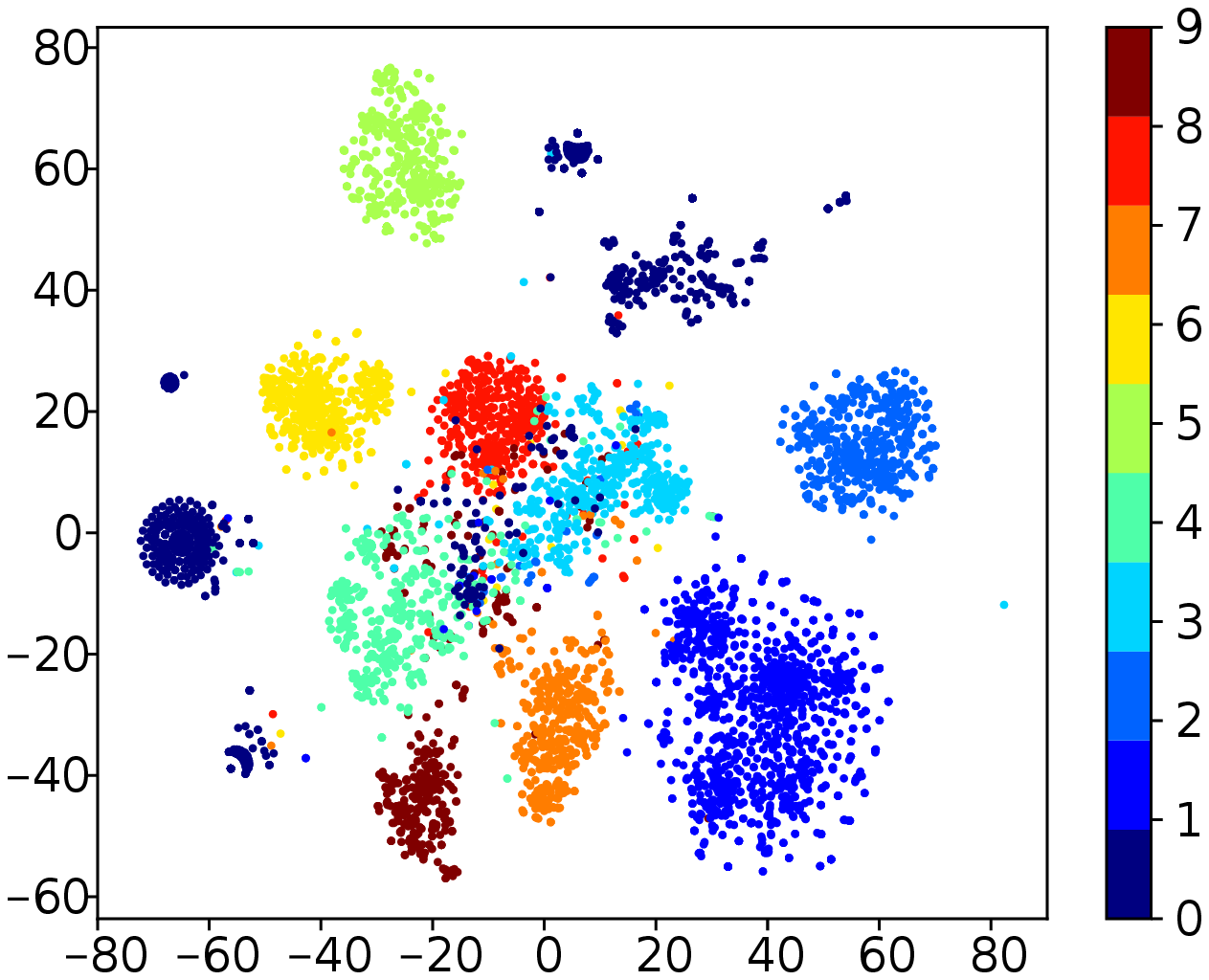} }\hfill
     \subfloat[BERT\label{BERT}]{\includegraphics[width=0.33\textwidth]{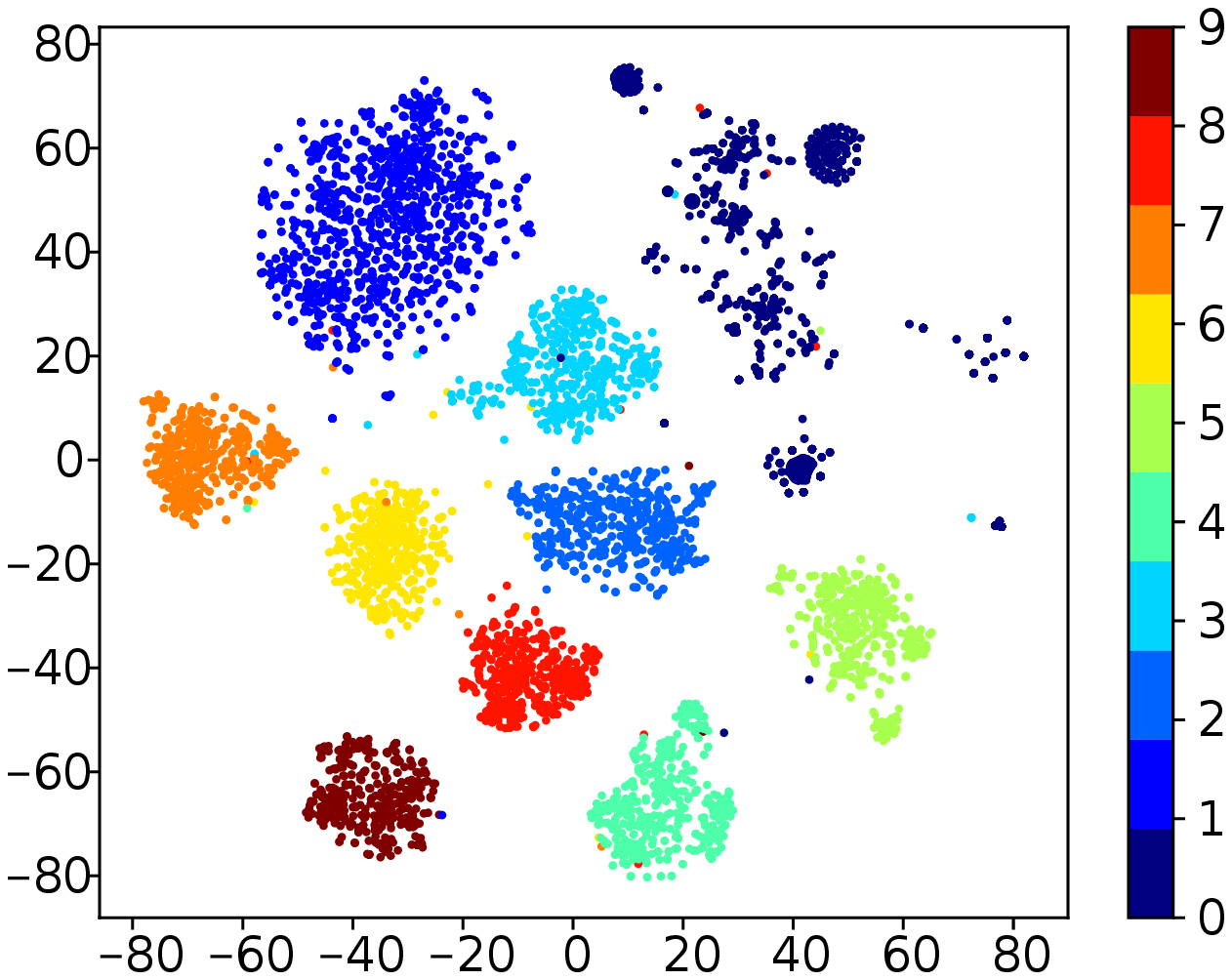}} \hfill \\
    \subfloat[AE\label{Autoencoder}]{\includegraphics[width=0.33\textwidth]{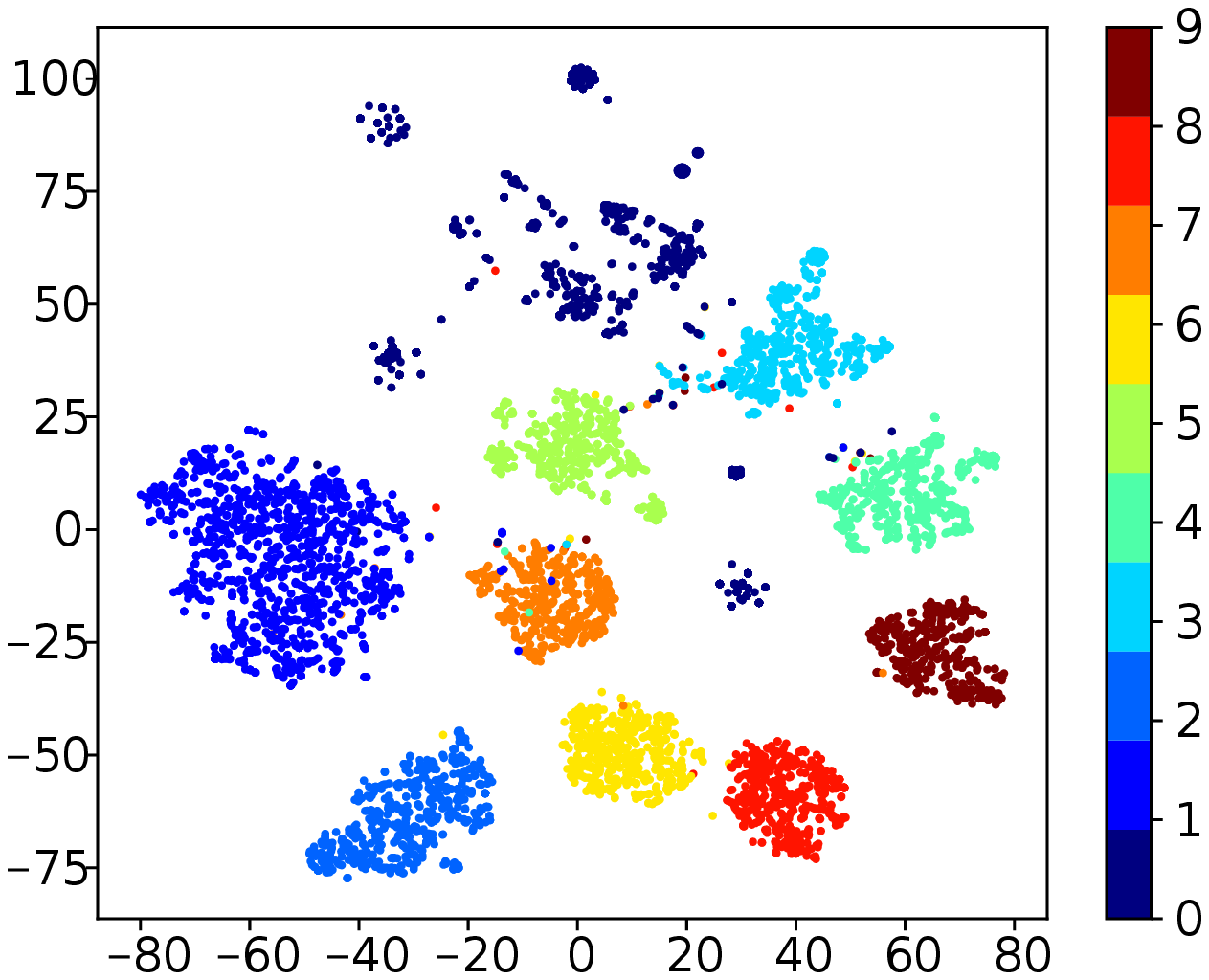}}\hfill
    \subfloat[STN-GAE\label{GAE}]{\includegraphics[width=0.33\textwidth]{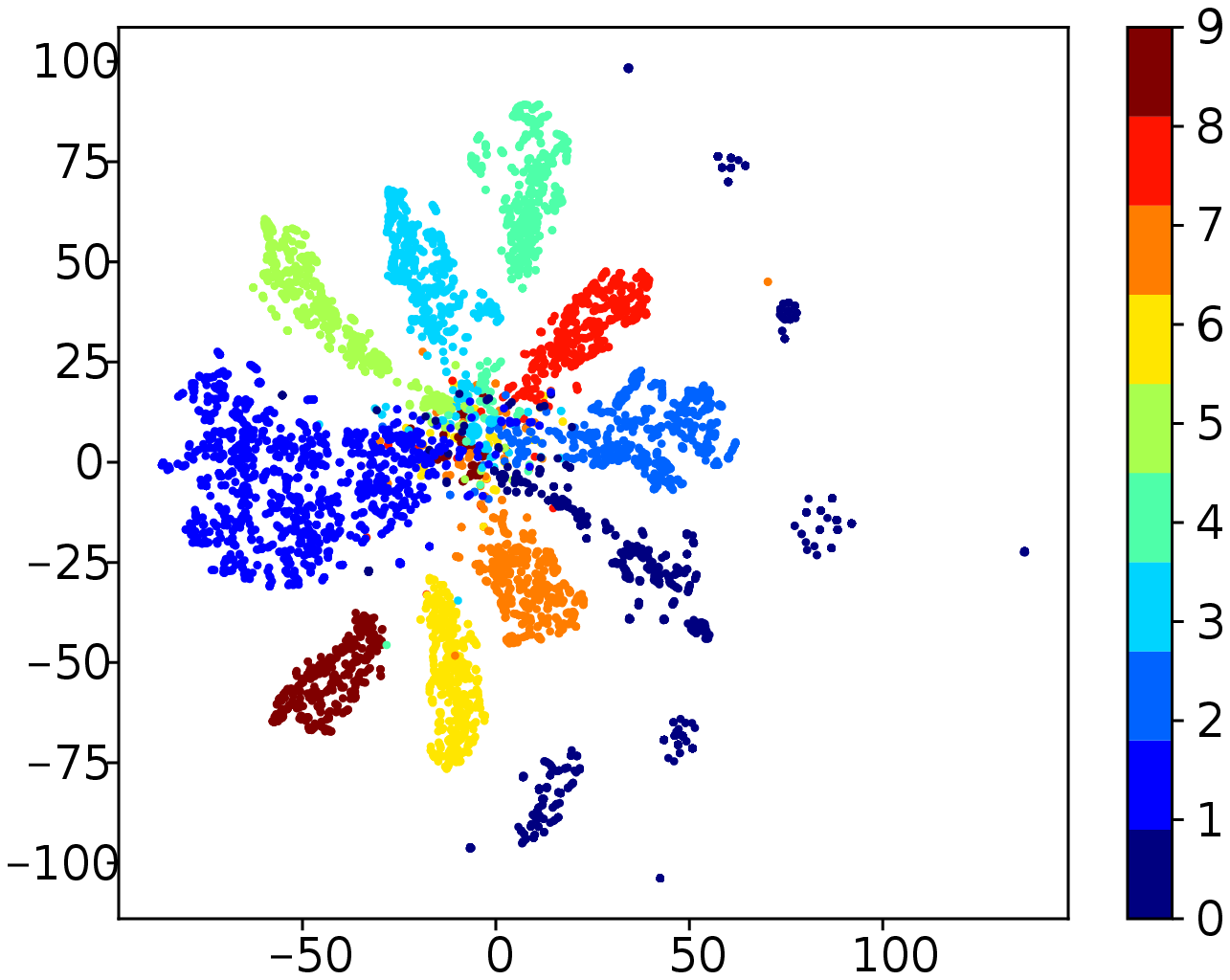}}\hfill
    \subfloat[SCA-AE\label{Soft}]{\includegraphics[width=0.33\textwidth]{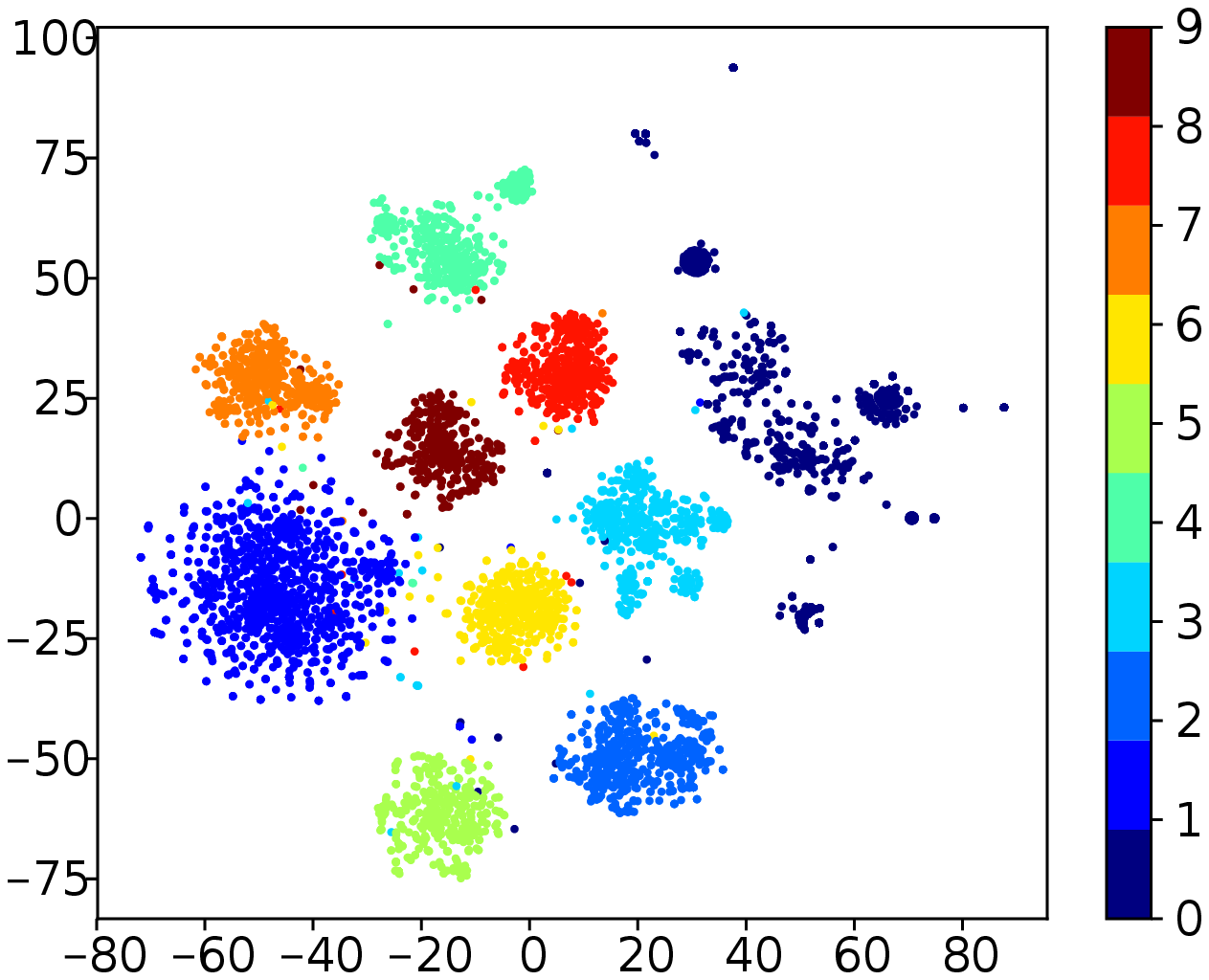}}
   
     \caption{t-SNE visualizations for clustering on Tweets dataset in (a) Bag of words space, (b) Word2vec space, (c) BERT embedding space, (d) Autoencoder hidden layer space, (e) Graph Autoencoder hidden layer space and (f) Autoencoder hidden layer space with soft cluster assignment. The true cluster labels are indicated using different colors.}
     \label{ModelPerformanceInEachClass}
\end{figure*}
Figure \ref{ModelPerformanceInEachClass} displays t-distributed stochastic neighbor embedding (t-SNE) visualizations for clustering on Tweets dataset in (a) Bag of words space, (b) Word2vec space, (c) BERT embedding space, (d) Autoencoder hidden layer space, (e) Graph Autoencoder hidden layer space and (f) Autoencoder hidden layer space with soft cluster assignment.
Figures \ref{BoW} and \ref{Word2vec} show that the samples are mixed together in a high-dimensional space without clear boundaries.
In figure \ref{BERT}, there are clear boundaries between nodes of different categories, but they are not concentrated enough and somewhat scattered, which proves that the BERT model has excellent representation capabilities.
Compared to figure \ref{BERT} BERT embedding space, in figure \ref{Soft}, the sample is more compact from the centroid. 
This is the feature of soft assignment, which makes the sample closer to the centroid and improves the NMI score.

\subsection{Parameters Analysis}
The model SCA-AE has achieved the best performance. 
We further analyze the relevant parameters to verify their impact on the model.
\subsubsection{Epochs of SCA-AE}
\label{EpochAnalysis}
\begin{figure}[t]
    \centering
    \includegraphics[width=125mm, scale=1.25 ]{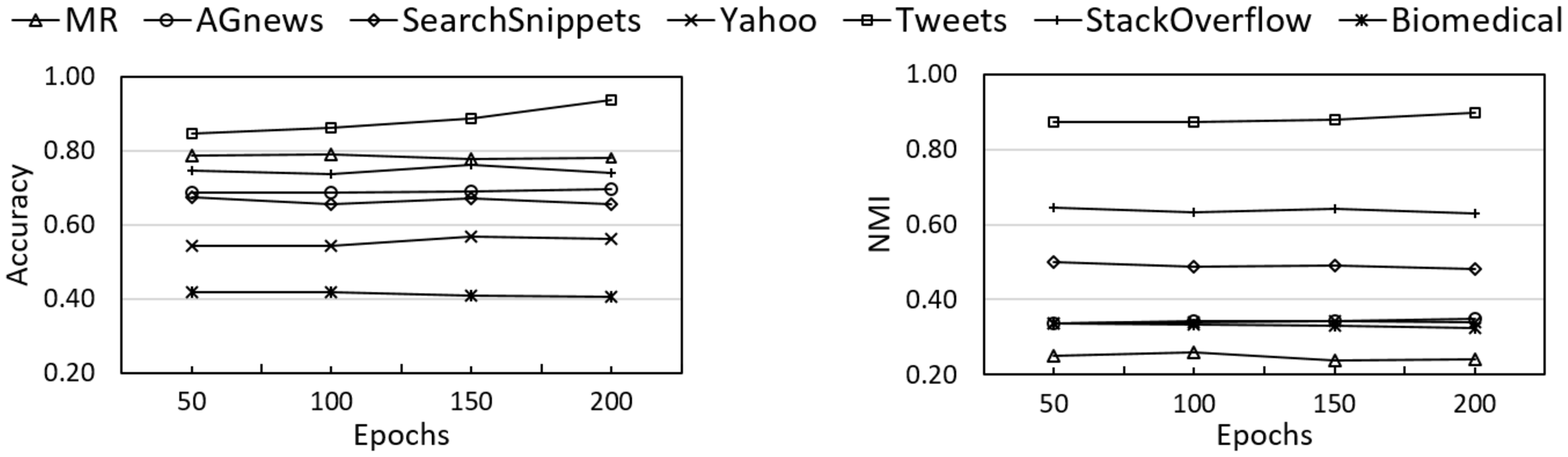}
    \caption{Epoch size comparison for SCA-AE.}
    \label{EpochsTest}
\end{figure}

In the soft cluster assignment method (SCA-AE), we vary the number of the epochs in autoencoder while keeping others unchanged.
Figure \ref{EpochsTest} shows the accuracy and NMI values of the seven datasets with different pre-trained epochs. 
With the increase of epochs in the autoencoder, the accuracy and NMI values of the tweet dataset have improved slightly, while other datasets have not changed much.
This demonstrates that the four-hidden-layer autoencoder is able to generate effective low-dimensional text representation in a few epochs.
\subsubsection{Autoencoder Layers of SCA-AE}  
\label{LayerAnalysis}
\begin{figure}[t]
    \centering
    \subfloat[Accuracy]{\label{ACCLayers}\includegraphics[width=0.5\textwidth]{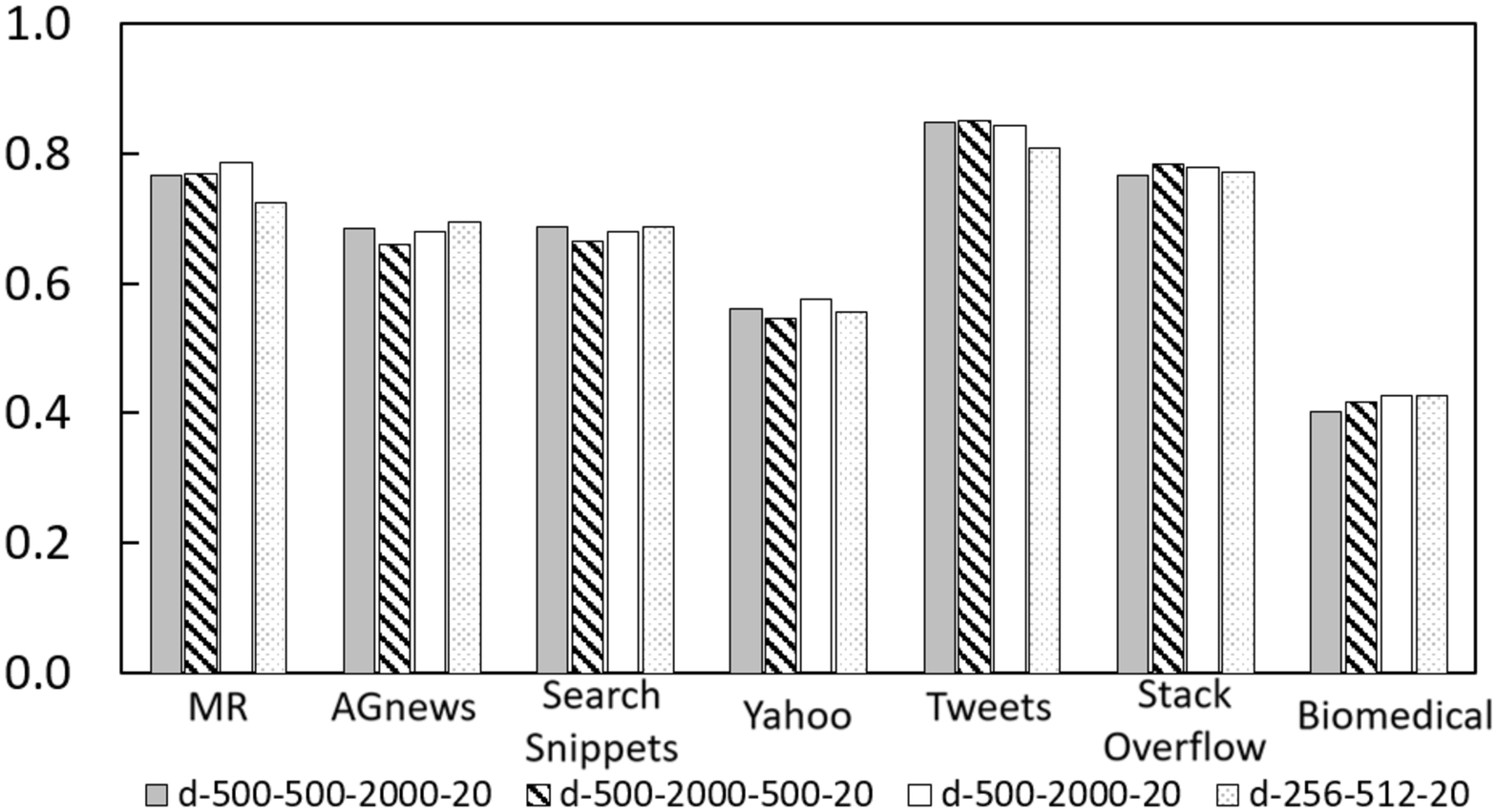}} 
    \subfloat[NMI]{\label{NMILayers}\includegraphics[width=0.5\textwidth]{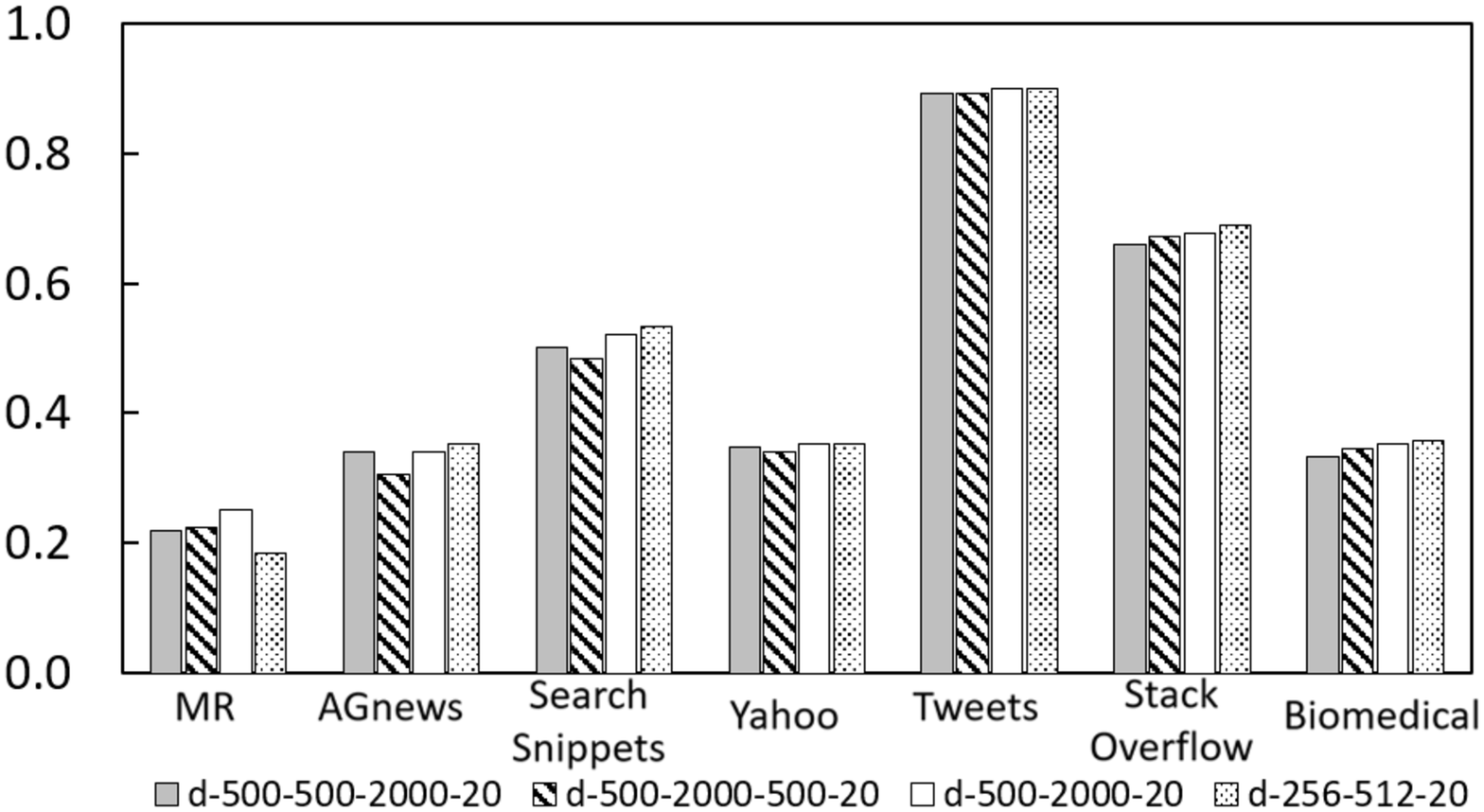}} 
    \caption{Hidden layer size comparison for SCA-AE.}
    \label{LayerTest}
\end{figure}

In the SCA-AE model, the hidden layer size of autoencoder is set to $d$:500:500:2000:20, which is the same as \cite{hadifar2019self}.
We change layer numbers and layer sizes of autoencoder hidden layers to verify the impact on the results.
$d$:500:2000:500:20 changes the order of hidden layers, $d$:500:2000:20 removes a hidden layer, and $d$:256:512:20 removes a hidden layer while reducing the number of neurons in each layer.
Figure \ref{LayerTest} shows the accuracy and NMI comparison based on the above hidden layer settings.
Although we have reduced the number of hidden layers or reduce the neuron number of the layer, the accuracy and NMI do not fluctuate much, so when using autoencoder for text representation dimensionality reduction, simply hidden layers and \yh{a small number of the neuron} can achieve effective results.

\section{Conclusions}
\label{conclusion}

In this work, we propose two methods to further enhance the short text representation learning based on the pre-trained text models to obtain the best clustering performance.
One is the integration of text structure information into the representation (STN-GAE), and the other is the soft cluster assignment constraint (SCA-AE).
\xy{We find that the BERT model produces effective text representations compared to traditional models (e.g., Word2vec, bag of words).}
Experimental results show that 
\xy{the integration of} text structure information into text representation (STN-GAE) has no positive impact on short text clustering, \xy{suggesting} that the nodes in the constructed short text network cannot obtain supplementary information from \xy{neighboring} nodes.
The proposed method, SCA-AE, further \xy{fine-tunes} the text representation and does improve the clustering performance. 

\section*{Acknowledgement}
This work was mainly supported by the Australian Research Council Linkage Project under Grant No. LP180100750.

\bibliographystyle{splncs04}

\bibliography{refer}

\end{document}